\documentclass{article}
\usepackage{spconf,amsmath,graphicx}

\usepackage[utf8]{inputenc} 
\usepackage[T1]{fontenc}    
\usepackage{hyperref}       
\usepackage{url}            
\usepackage{booktabs}       
\usepackage{amsfonts}       
\usepackage{nicefrac}       
\usepackage{microtype}      
\usepackage{lipsum}		
\usepackage{graphicx}
\usepackage{doi}

\usepackage{algorithm}
\usepackage{algpseudocode}
\usepackage{array}
\usepackage{mathtools}
\usepackage{multirow}
\usepackage{setspace}

\newcolumntype{L}[1]{>{\raggedright\let\newline\\\arraybackslash\hspace{0pt}}m{#1}}
\newcolumntype{C}[1]{>{\centering\let\newline\\\arraybackslash\hspace{0pt}}m{#1}}
\newcolumntype{R}[1]{>{\raggedleft\let\newline\\\arraybackslash\hspace{0pt}}m{#1}}

\title{\rlap{Amicable Aid: Perturbing Images to Improve Classification Performance}}
\name{Juyeop Kim, Jun-Ho Choi, Soobeom Jang, Jong-Seok Lee}
\address{Yonsei University, Korea}

\begin{document}
\ninept
\maketitle
\begin{abstract}
While adversarial perturbation of images to attack deep image classification models pose serious security concerns in practice, this paper suggests a novel paradigm where the concept of image perturbation can benefit classification performance, which we call \emph{amicable aid}.
We show that by taking the opposite search direction of perturbation, an image can be modified to yield higher classification confidence and even a misclassified image can be made correctly classified.
This can be also achieved with a large amount of perturbation by which the image is made unrecognizable by human eyes.
The mechanism of the amicable aid is explained in the viewpoint of the underlying natural image manifold.
Furthermore, we investigate the \emph{universal amicable aid}, i.e., a fixed perturbation can be applied to multiple images to improve their classification results.
While it is challenging to find such perturbations, we show that making the decision boundary as perpendicular to the image manifold as possible via training with modified data is effective to obtain a model for which universal amicable perturbations are more easily found.
\end{abstract}
\section{Introduction}
\label{sec:intro}

Deep learning technologies have been widely expanded to various applications thanks to the abundance of data and computational resources.
However, recent studies have shown that deep neural networks are highly vulnerable to adversarial attacks, which fool the target deep model by adding imperceptible noise to input images.
For instance, adversarial attacks can make image classification models misclassify a given image~\cite{su2018robustness}.
Such vulnerability to adversarial attacks raises significant security concerns in real-world applications~\cite{goodfellow2015explaining}.

An attack can be defined as adding perturbation $\delta$ to image $x$ to mislead an image classification model.
There are two natures that make this an ``attack''~\cite{hwang2021just}.
First, $\delta$ should be obtained so that the model fails to recognize the image as a ground-truth class $y$ (i.e., deterioration).
Second, $\delta$ should be small enough to conceal the fact that the original image has been manipulated (i.e., inconspicuousness).

\begin{figure}[!t]
\centering
\small
    \hspace{-2mm}
    \includegraphics[width=0.99\linewidth]{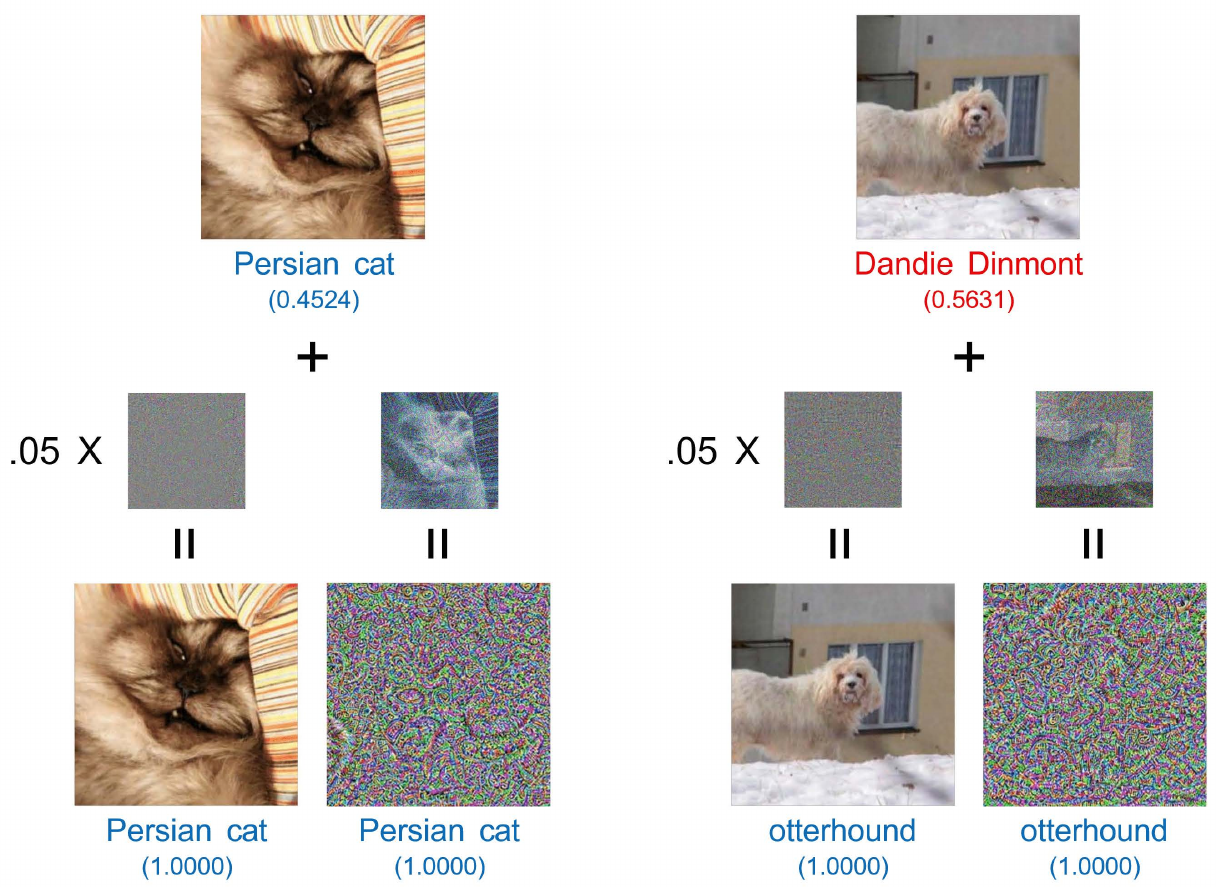} \\
    \hspace{3.5mm} (a) \hspace{12.5mm} (b) \hspace{23mm} (c) \hspace{12.5mm} (d)
    \caption{
        Concept of the proposed amicable aid.
        \textbf{Left}: The original image is correctly classified as ``Persian cat'' by the trained VGG16 \cite{simonyan2015very} model with confidence of 0.4524.
        (a) By adding a small perturbation computed by our method, called weak aid, the confidence can be maximized.
        (b) Confidence boosting can be achieved even with a large amount of perturbation that completely destroys the original content, called strong aid.
        \textbf{Right}: The original image (``otterhound'') is misclassified by VGG16 as ``Dandie Dinmont.''
        (c) By the weak aid, the image can be correctly classified with maximum confidence.
        (d) The strong aid also can change the image in a way that it is correctly classified with maximum confidence.
    }
    \label{fig: amicable aid}
\end{figure}

A question that motivated our work in this paper is as follows:
\emph{If it is possible to find a perturbation that can deteriorate classification performance, will it be also possible to find a perturbation that can \textbf{improve} classification performance?}
We find that the answer is yes and doing so is surprisingly effective.
Adding an imperceptible perturbation pattern to an image can not only increase the confidence score when the original classification result is correct (Fig.~\ref{fig: amicable aid}a), but also make the classification result correct even if the model misclassifies the original image (Fig.~\ref{fig: amicable aid}c).
More surprisingly, it is also possible to add a significant amount of perturbation to increase confidence or fix the classification result so that the original content is completely destroyed (Figs.~\ref{fig: amicable aid}b and \ref{fig: amicable aid}d).
We call this phenomenon \textbf{amicable aid} as an opposite concept to adversarial attack.
We can rethink the role of image perturbations that have been used for adversarial attacks based on the two natures, i.e., deterioration and inconspicuousness.
Our contributions can be summarized as follows.

\noindent $\bullet$ We define the notion of amicable aid and show that the amicable aid can be performed with high success rates by iterative gradient-based search. 
We also define two types of amicable aid: the weak amicable aid with imperceptible perturbations and the strong amicable aid with perceptible perturbations destroying the original image content.

\noindent $\bullet$ We explain the mechanism of the amicable aid in the viewpoint of image manifold and decision boundary.

\noindent $\bullet$ We examine the feasibility of finding universal perturbations of the amicable aid.
We show that while finding a universal perturbation is challenging, the challenge can be resolved by making the decision boundaries more perpendicular to the data manifold.

\section{Related work}

Recent studies have shown that many deep learning-based image classification models are highly vulnerable to malicious image manipulation, e.g., adding a hardly imperceptible perturbation~\cite{szegedy2013intriguing}, which is known as the adversarial attack.
As notable methods, the fast gradient sign method (FGSM) and its iterative version (called I-FGSM) were developed~\cite{goodfellow2015explaining,kurakin2016adversarial}.
Several other adversarial attack techniques have been proposed with various perspectives and objectives \cite{carlini2017evaluating,luo2018towards,madry2017towards,tramer2018ensemble,modas2019sparsefool,wong2020fast,croce2020reliable}.
Existence of an image-agnostic universal perturbation has been also shown \cite{moosavi2017universal}, i.e., a single perturbation can attack multiple images.

A plausible explanation of adversarial attacks is the manifold theory.
While natural images lie on a low-dimensional data manifold, adding an adversarial perturbation to an image causes the image to leave the manifold and also cross the decision boundary.
In this sense, attacks adding noise-like perturbations can be called off-manifold attacks, whereas on-manifold adversarial samples remaining on the manifold can be also found~\cite{stutz2019disentangling,song18constructing}.
Finding off-manifold samples becomes easy when the decision boundary lies close to the manifold \cite{tanay2016boundary}.

While adversarial attacks are undesirable threats for classifiers in general, there have been a few attempts to exploit attacks for specific purposes recently, e.g. to deal with model stealing~\cite{lukas2019deep}, to improve generalization performance via regularization~\cite{xie2020adversarial}, to explain learned representations~\cite{jalwana2020attack}, and to explore noisy images that are classified with high confidence~\cite{deep2015nguyen,InkawhichLWICC20,zhao2021on}.
To the best of our knowledge, \cite{huang2021unlearnable,Pestana_2021_CVPR,salman2021unadversarial} are the only attempts to search perturbations minimizing the classification error.
In \cite{huang2021unlearnable}, a method for privacy protection was proposed by making an image unlearnable with a perturbation during model training.
This method and ours commonly set the direction for searching perturbations to the opposite of that of adversarial attacks.
However, the former aims to fool the training process, whereas our work focuses on understanding and analyzing the characteristics and mechanism of error-minimizing perturbations in the viewpoint of improving classification performance.
In \cite{Pestana_2021_CVPR,salman2021unadversarial}, perturbations are also found from the opposite direction to adversarial attacks.
However, the works show the feasibility of the idea only on a limited, specific dataset (mesh data \cite{Pestana_2021_CVPR,salman2021unadversarial}) or a specific perturbation type (small patches \cite{salman2021unadversarial}) without analysis.
Our work considers more general image classification datasets and provides deeper understanding of the amicable aid.
Furthermore, we also present an image modification method that facilitates optimization of universal error-minimizing perturbations.

\begin{table}[!t]
\centering
{\small
\caption{
Classification accuracy (with average confidence for the true class labels) for the original images, weakly aided images, and strongly aided images for CIFAR-100.
}\label{tab:aid_on_cifar100}
\vspace{2mm}
\begin{tabular}{rrrrr}

    \toprule
    \toprule
    
    \multicolumn{1}{c}{} &
    \multicolumn{1}{c}{} &
    \multicolumn{1}{c}{Original} &
    \multicolumn{1}{c}{Weak aid} &
    \multicolumn{1}{c}{Strong aid} \\
    
    \midrule
    
    \multirow{2}{*}{VGG16} & &
       72.30\% & 100.0\% & 98.37\% \\
    && (.7090) & (1.000) & (.9767) \\
    \midrule
    \multirow{2}{*}{ResNet50} & &
       78.79\% & 100.0\% & 97.45\% \\
    && (.7575) & (1.000) & (.9677) \\
    \midrule
    \multirow{2}{*}{MobileNetV2} & &
       68.89\% & 100.0\% & 99.48\% \\
    && (.6182) & (1.000) & (.9800) \\
    
    \bottomrule
    \bottomrule

\end{tabular}
}
\end{table}

\begin{table}[!t]
\centering
{\small
\caption{
Classification accuracy (with average confidence for the true class labels) for the original images, weakly aided images, and strongly aided images for ImageNet.
}\label{tab:aid_on_imagenet}
\vspace{2mm}
\begin{tabular}{rrrrr}

    \toprule
    \toprule
    
    \multicolumn{1}{c}{} &
    \multicolumn{1}{c}{} &
    \multicolumn{1}{c}{Original} &
    \multicolumn{1}{c}{Weak aid} &
    \multicolumn{1}{c}{Strong aid} \\
    
    \midrule
    
    \multirow{2}{*}{VGG16} & &
       71.59\% & 100.0\% & 77.59\% \\
    && (.6344) & (1.000) & (.7448) \\
    \midrule
    \multirow{2}{*}{ResNet50} & &
       76.15\% & 100.0\% & 87.62\% \\
    && (.6941) & (1.000) & (.8456) \\
    \midrule
    \multirow{2}{*}{MobileNetV2} & &
       71.87\% & 100.0\% & 91.53\% \\
    && (.6358) & (1.000) & (.8933) \\
    
    \bottomrule
    \bottomrule

\end{tabular}
}
\end{table}

\section{Amicable aid}
\label{sec:amicable_aid}

Given an image classification model with weight parameters $\theta$ and an image $x$ with the true class label $y$, performing the amicable aid can be formulated as an optimization problem as follows:
\begin{align}
\small
    \min_\delta J(x+\delta, y, \theta) 
    \mathrm{~~~such~that~~}||\delta||_p \leq \epsilon,
\end{align}
where $\delta$ is the perturbation, $J$ is the cost function (e.g., cross-entropy), and $\epsilon$ is a hyperparameter. In other words, we want to find $\delta$ that minimizes the cost function while its $p$-norm is bounded by $\epsilon$.
Note that this formulation is the opposite to that for adversarial attacks where minimization is replaced by maximization.

The above optimization problem can be solved by adopting the formulations of gradient-based adversarial attack methods.
For instance, the idea of FGSM~\cite{goodfellow2015explaining}, which is one of the most popular and strongest attacks, can be used for $p=\infty$: 
\begin{align}
\small
    \delta = -\epsilon \cdot \mathrm{sign}\big({\nabla}_{x}J(x, y, \theta)\big).
\label{eq:fgsm_aid}
\end{align}
Note that the negative signed gradient is used here as opposed to the FGSM attack.
It is known that FGSM can be made more powerful using an iterative approach, I-FGSM~\cite{kurakin2016adversarial}.
Similarly, an iterative version of (\ref{eq:fgsm_aid}) can be formulated as follows.
\begin{align}
\small
    x_0 &= x, \\
    \tilde{x}_{n+1} &= \mathrm{clip}_{0,1}\big(x_n - \frac{\epsilon}{N}\mathrm{sign}({\nabla}_{x}J(x_{n}, y, \theta))\big), \label{eq:ifgsm_aid0}\\
    x_{n+1} &= \mathrm{clip}_{-\epsilon,\epsilon}(\tilde{x}_{n+1}-x)+x,
    \label{eq:ifgsm_aid}
\end{align}
where $N$ is the number of iterations and
    $\mathrm{clip}_{a,b}(z) = \min(\max(z,a),b)$.

As shown in Fig.~\ref{fig: amicable aid}, $\epsilon$ can be controlled so that the perturbation becomes either imperceptible with a small $\epsilon$ ($\epsilon=.05$ in Fig.~\ref{fig: amicable aid}a and \ref{fig: amicable aid}c) or perceptible with a large $\epsilon$ ($\epsilon=10$ in Fig.~\ref{fig: amicable aid}b and \ref{fig: amicable aid}d), which we call \textbf{weak aid} and \textbf{strong aid}, respectively.

We apply the weak and strong aids to the test images of CIFAR-100~\cite{krizhevsky2009learning} and the validation set of ImageNet~\cite{ILSVRC15}.
Three models, namely, VGG16 \cite{simonyan2015very}, ResNet50 \cite{he2016deep}, and MobileNetV2 \cite{sandler2018mobilenetv2}, are employed.
For ImageNet, we use the pre-trained models provided by PyTorch.
For CIFAR-100, we train the models to minimize the cross-entropy loss using the SGD optimizer for 200 epochs with an initial learning rate of 0.1, a momentum parameter of 0.9, and a weight decay parameter of .0005.
We use $\epsilon=.05$ for the weak aid, which is small enough to hide the perturbations, and $\epsilon=10$ for the strong aid, which is large enough to introduce large distortions in the images\footnote{Note that the maximum amount of the added perturbation at each iteration is $\epsilon/N$ in (4), which is 10/100=0.1 for the strong aid.}.
The number of iterations ($N$) is set to 100.
The results are shown in Tables~\ref{tab:aid_on_cifar100} and \ref{tab:aid_on_imagenet}.
It is observed that the aids significantly improve the classification performance, showing near-perfect or perfect classification accuracies and confidence scores.
Finding perturbations for the strong aid is slightly more difficult compared to the weak aid due to the requirement of large perturbation changes at each search iteration.

\begin{figure}[!t]
\small
  \centering
  \includegraphics[width=.5\linewidth]{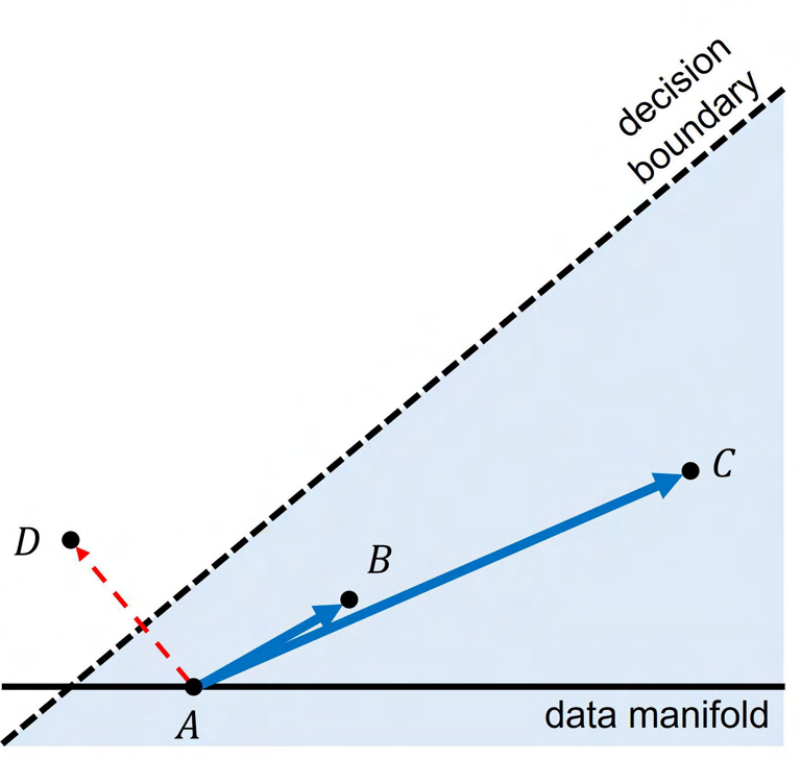}
  \caption{\label{fig:mechanism}
  Illustration of the process of the amicable aid. Point $A$ indicates the original image residing on the manifold. The blue solid arrows mean the modification process of the original image by the amicable aid, while the red dashed arrow means the modification process by the attack. Points $B$ and $C$ correspond to a weakly and a strongly aided images, respectively. Both attack and aid make the original image leave the manifold, but the aided images still remain inside the decision region (shaded region) while the attacked image does not.
  }
\end{figure}

\begin{figure}[!t]
\small
  \centering
  \includegraphics[width=.6\linewidth]{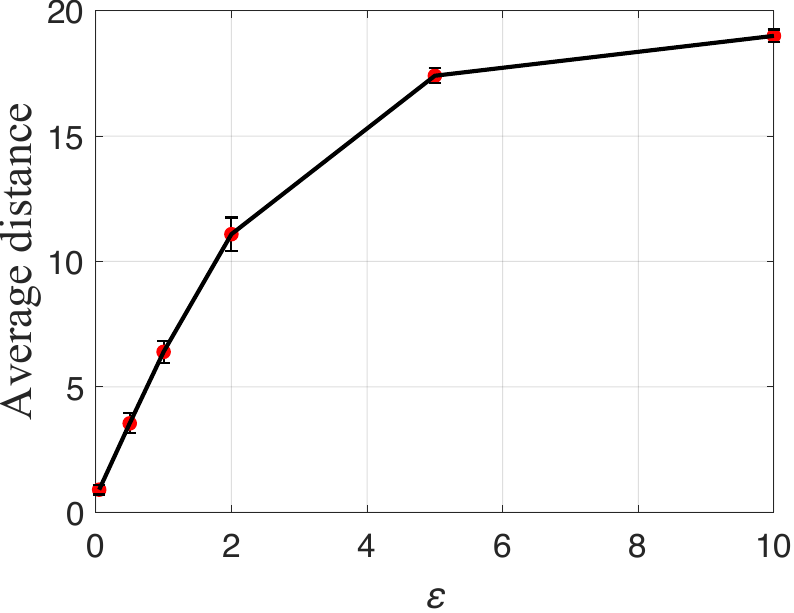}
  \caption{\label{fig:kNN_distance}
  Average distance from the aided images to the approximated manifold with respect to the value of $\epsilon$.
  }
\end{figure}

\section{Mechanism}

In the manifold-based explanation for adversarial attacks, ordinary adversarial attacks that modify images in unnatural ways cause the images leave the manifold \cite{stutz2019disentangling}.
The mechanism of the amicable aid can be also explained in the viewpoint of the manifold and decision boundary, as illustrated in Fig.~\ref{fig:mechanism}.
The original image ($A$), which is inside the decision boundary and thus correctly classified, lies on the manifold. By an off-manifold attack, it leaves the manifold and also goes beyond the decision boundary so that the classification result becomes wrong ($D$).
Amicable aids also make the image leave the manifold, but the perturbed images ($B$ and $C$) still stay within the decision region.
The strongly aided image ($C$) is located farther from the manifold due to a larger amount of perturbation by a large value of $\epsilon$ than the weakly aided image ($B$) using a small value of $\epsilon$.

To support this explanation, we check the distance of aided images from the manifold.
The true manifold is not known, thus it is approximated by the nearest neighbor-based method described in \cite{stutz2019disentangling}.
Fig.~\ref{fig:kNN_distance} shows the average distance of the aided images to the approximated manifold with respect to the value of $\epsilon$ for the test images of CIFAR-100 when VGG16 is employed.
It is clearly observed that the distance to the manifold increases as $\epsilon$ increases.

\begin{figure}[!t]
\centering
\small
    \begin{tabular}{cc}
    \multicolumn{2}{c}{\includegraphics[width=0.8\linewidth]{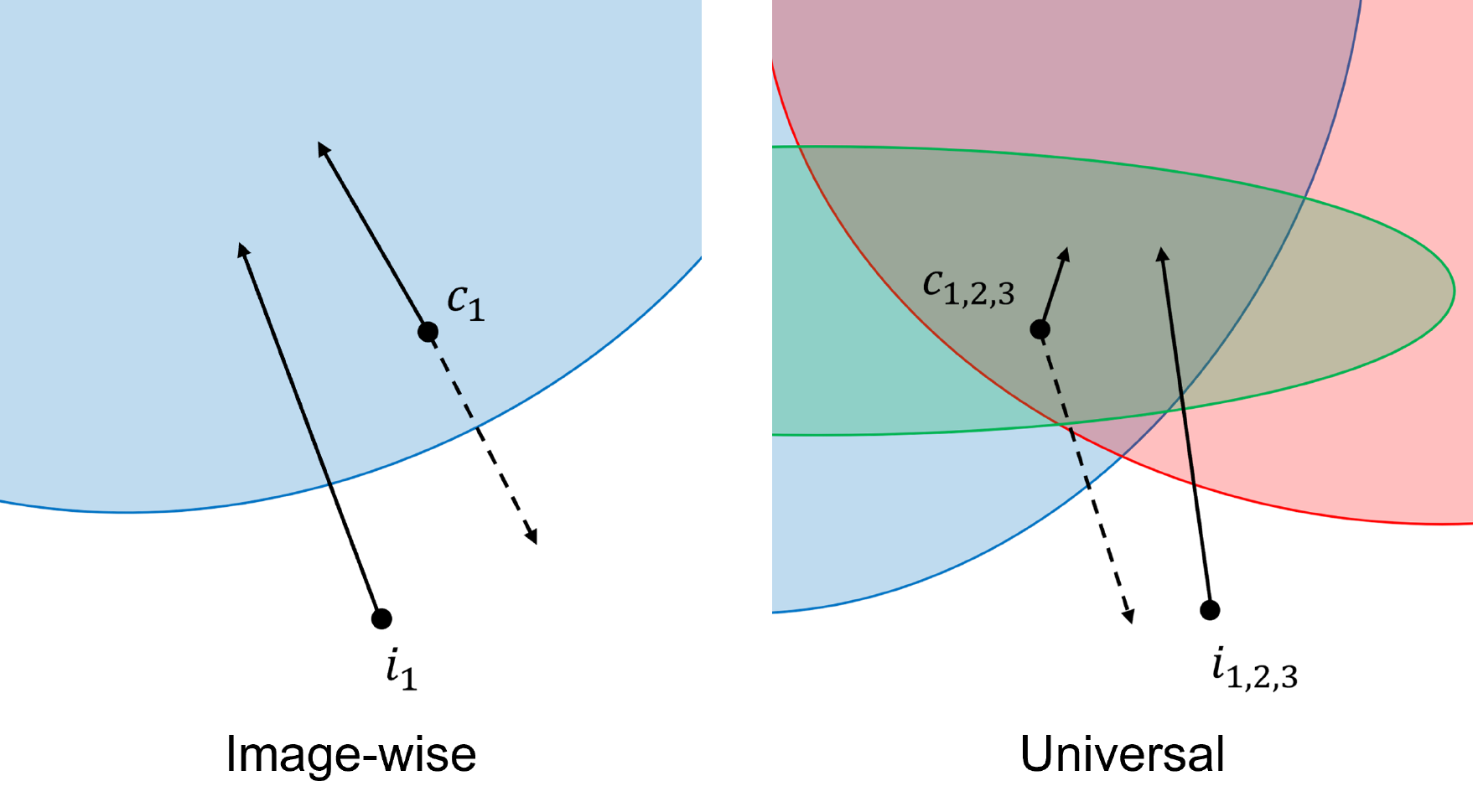}} \vspace{-2mm} \\
    \end{tabular}
    \caption{\label{fig:image_wise_vs_universal}
        Illustration of the image-wise aid and the universal aid (solid lines).
        A shaded region represents the decision region where a correct classification result is obtained.
        The image-wise attack and the universal attack are also shown (dotted lines) for comparison. 
        \textbf{Left}: An original image ($c_1$) that is correctly classified is moved within the decision region by the amicable aid.
        An image ($i_1$) that is misclassified is moved into the decision region by the amicable aid.
        \textbf{Right}: Three correctly classified images ($c_1$, $c_2$, and $c_3$) are super-imposed and their corresponding decision regions are shown with different colors.
        The universal amicable aid (solid line) needs to move the three images within the intersection of the three decision regions.
        Three misclassifed images ($i_1$, $i_2$, and $i_3$) also need to be sent inside the intersection by the universal aid.
        On the other hand, the universal attack (dotted line) needs to move three images outside their decision regions, where the region of the valid universal attack is far wider than that of the universal aid.
    }
\end{figure}

\begin{figure}[!t]
\centering
\small
    \begin{tabular}{cc}
        \multicolumn{2}{c}{\includegraphics[width=0.8\linewidth]{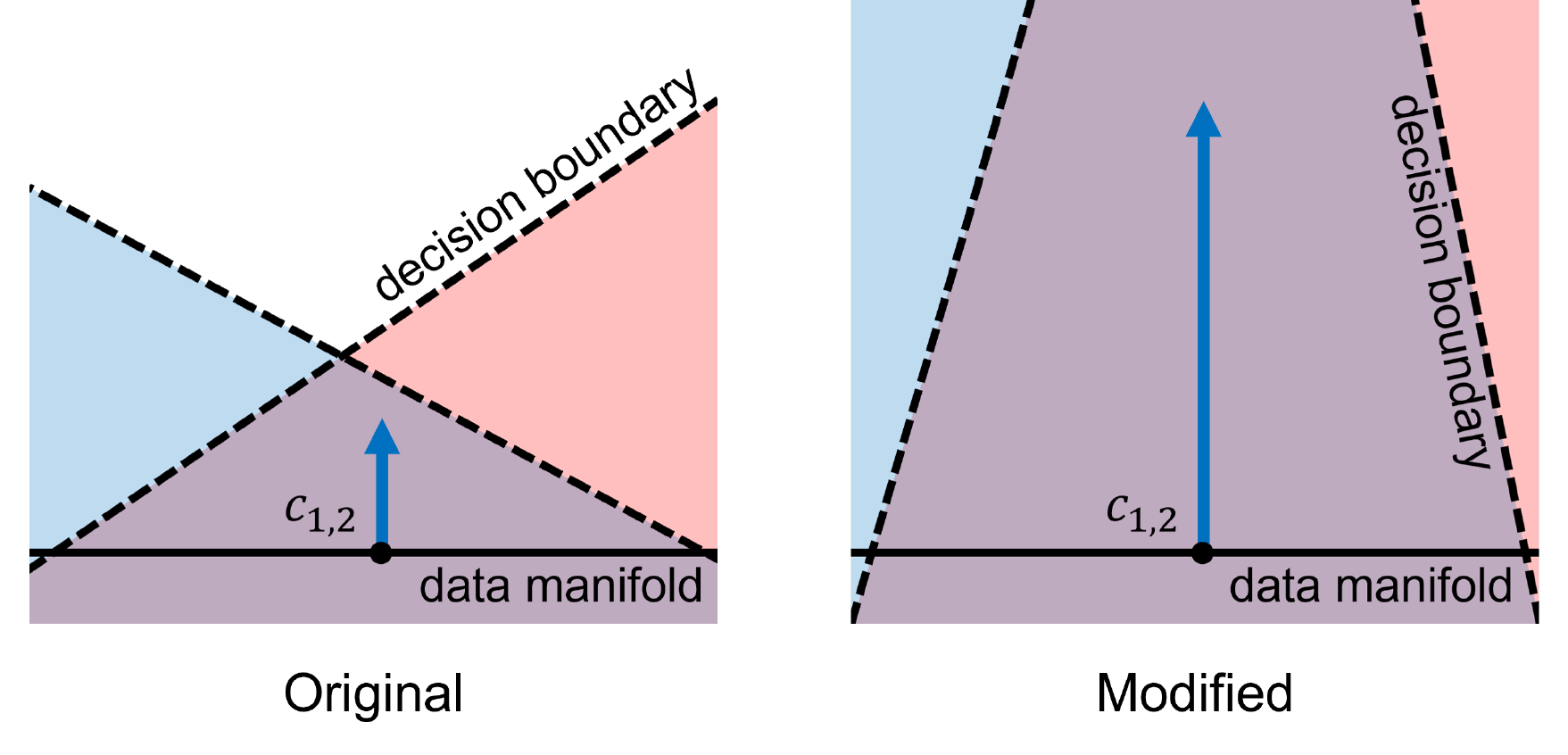}} \vspace{-2mm}
    \end{tabular}
    \caption{\label{fig:original_vs_modified}
        Illustration of the universal aid for an ordinary model and a model trained on a modified dataset.
        Two correctly classified images ($c_1$ and $c_2$) are super-imposed and their corresponding decision regions are shown with different colors.
        A straight line represents the data manifold where the natural images reside while dotted lines represent the decision boundaries.
        \textbf{Left}: A universal aid cannot make the two images go beyond the intersection of the decision boundaries.
        \textbf{Right}: A dataset containing modified images is used to train the model so that the decision boundaries have larger angles with the data manifold, which expands the intersection of the decision regions and thus allows the universal aid to be successful with a higher strength.
    }
\end{figure}

\begin{figure*}[!t]
    \small
    \centering
    \begin{tabular}{ccc}
        \multicolumn{3}{c}{
        \hspace{5mm}\includegraphics[width=.45\linewidth]{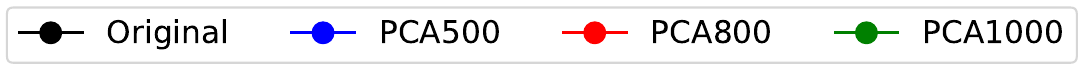}
        } \\
        \includegraphics[width=.25\linewidth]{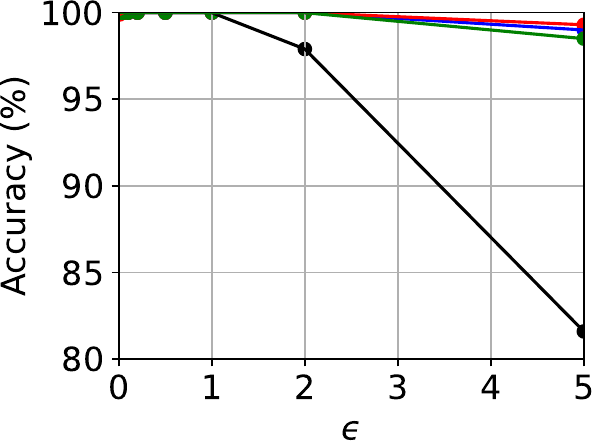} &
        \includegraphics[width=.24\linewidth]{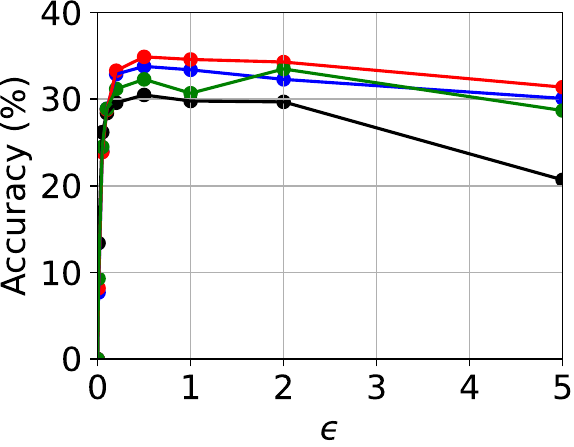} &
        \includegraphics[width=.24\linewidth]{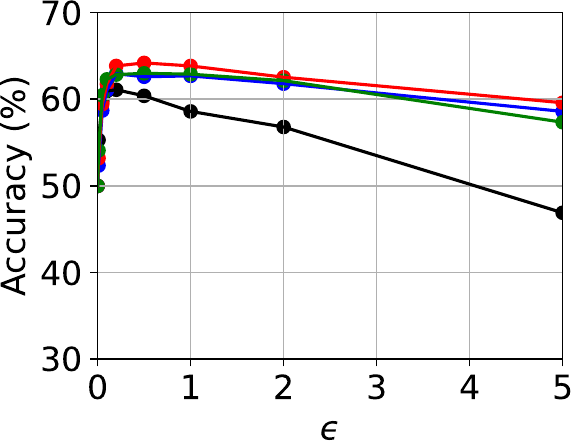} \\
        \hspace{8mm}(a) & \hspace{6.5mm}(b) & \hspace{7.5mm}(c) \vspace{-1mm}
    \end{tabular}
    \caption{\label{fig:universal}
        Classification accuracy of the original model (black) and models trained using modified images  (blue, red, and green). The amicable perturbations are found for and applied to (a) \texttt{correct}, (b) \texttt{incorrect}, and (c) \texttt{mix}.
    }
\end{figure*}

\begin{algorithm}[!t]
\small
\caption{Model training using samples modified by PCA}
\textbf{Input: }Training images $X=\{x|x\in[0, 1]^{D}\}$ and its covariance matrix $C$, training labels $Y$, and mean image $\bar{x}$ \\
\textbf{Parameters: } Dimension of the PCA subspace $d$, dimension of additive noise $m$, noise scale $c > 0$, and desired model structure $s$ \\
\textbf{Output: } Trained model with modified dataset. \\
\begin{spacing}{0.8}
\begin{algorithmic}[1]
\setstretch{1.1}
\vspace{-4mm}
    \State Define a set of modified images $X^\prime \gets \{ \}$.
    
    \State Perform spectral decomposition on $C$.\newline
    \hspace*{1em} $C = \sum_{i=1}^{D} \lambda_{i}v_{i}v_{i}^{\top}$, ~ $\lambda_{1} \geq \lambda_{2} \geq \dots \geq \lambda_{D}$,\newline
    where $\lambda_i$ and $v_i$ are an eigenvalue and the corresponding eigenvector of $C$, respectively.
    
    \For {$x \in X$}
    \State Project $x-\bar{x}$ on the eigenvector space.
    \newline
    \hspace*{3em} $a \gets V^{\top}\big(x-\bar{x}\big)$, ~ $V = \big[v_{1}, v_{2}, \ldots, v_{D}\big]$.
    
    \State Add noise to obtain $a^\prime = \big[ a^{\prime}_{1}, a^{\prime}_{2}, \ldots, a^{\prime}_{D} \big]$.
    \newline
    \hspace*{3em} $a^\prime_{i} =
    \begin{cases}
    a_{i}, & 1\leq i \leq d \\
    c \sqrt{\lambda_{i}} \xi_{i}, & d+1 \leq i \leq d+m \\
    0, & \text{otherwise},
    \end{cases}
    $ \newline
    \hspace*{1.2em} where $\xi_{i} \sim \mathcal{N}(0, 1)$. 
    
    \State Reconstruct modified image $x^\prime$. 
    \newline
    \hspace*{3em} $x^\prime \gets V a^\prime + \bar{x}$. 
    
    \State Add $x^\prime$ to $X^\prime$.
    
    \EndFor
    
    \State Train $s$ with $X^\prime$ to obtain the trained model.
\vspace{1mm}
\end{algorithmic}
\end{spacing}
\label{alg:pca}
\end{algorithm}

\section{Universal amicable aid}

In adversarial attacks, it is possible to find an image-agnostic perturbation that affects multiple images to make a given model misclassify them, which is known as the universal attack~\cite{moosavi2017universal}.
Inspired by this idea, the concept of \emph{universal amicable aid} can be proposed.
The objective of the universal amicable aid is to find an image-agnostic perturbation that can aid multiple images to obtain correct classification results.
Basically, this can be achieved by using the average gradient over multiple images in (\ref{eq:ifgsm_aid0}).
However, this is expected to be a more challenging problem than the universal attack as illustrated in Fig.~\ref{fig:image_wise_vs_universal}; multiple images are moved by the same direction and amount by the universal perturbation, and all the resulting images need to be inside the respective correct decision regions simultaneously.

Our idea to overcome the challenge is illustrated in Fig.~\ref{fig:original_vs_modified}.
If we can make the decision boundaries perpendicular to the data manifold, a large number of (correctly classified) images can leave the data manifold by a large distance while they still remain inside their decision regions.
To realize this, we adopt the data modification method presented in \cite{Li2020DefenseAA}.
In this method, the data for model training are modified to contain noise in the normal direction of the data manifold by performing principal component analysis (PCA) and then adding noise, so that the model learns decision boundaries perpendicular to the data manifold.
Algorithm \ref{alg:pca} summarizes the procedure of model training using the modified dataset by PCA.

To examine the efficacy of the universal amicable aid, we form three sets of images, which are referred to as \texttt{correct}, \texttt{incorrect}, and \texttt{mix}, respectively.
\texttt{correct} and \texttt{incorrect} consists of 1,000 images from the test set of CIFAR-100 that are originally classified correctly and incorrectly, respectively.
\texttt{mix} is the union of \texttt{correct} and \texttt{incorrect}.
Each of the three sets is used to find a universal amicable perturbation, respectively.
Then, the found universal perturbations are applied to themselves to evaluate performance of the universal amicable perturbations.

Fig.~\ref{fig:universal} shows the performance of our method for ResNet50 (similar trends are also obtained for VGG16 and MobileNetV2, which are omitted due to page limit).
Here, both hyperparameters $m$ and $c$ are set to 10, and $d$ varies among 500, 800, and 1000 (noted as PCA500, PCA800, and PCA1000 in the figure, respectively).
Other training hyperparameters are set as in Section~\ref{sec:amicable_aid}.
Note that the default accuracy without aid is 100\%, 0\%, and 50\% for \texttt{correct}, \texttt{incorrect}, and \texttt{mix}, respectively.
It is clearly observed that the models trained with the modified images show higher accuracy than the original model especially for large values of $\epsilon$, demonstrating the improvement by our method in finding universal amicable perturbations.

\section{Conclusion}

We have introduced the new concept of the amicable aid that adds a perturbation to a given image in order to improve the classification performance for the image.
We demonstrated that not only the weak aid with imperceptible perturbations but also the strong aid with perceptible perturbations are feasible.
Furthermore, we showed that it is possible to obtain high-quality universal amicable perturbations by controlling the geometry of the decision boundary during training.

This paper presented the first study on the amicable aid, and opens new research directions for which much work needs to be performed in the future.
(1) We presented one way to generate amicable perturbations, i.e., gradient-based off-manifold perturbations, but we expect that there are several other ways, e.g., on-manifold perturbations \cite{shamsabadi20colorfool,zhao20towards}, weight perturbations \cite{bai21targeted}, etc.
(2) This paper focused on fundamental characteristics of the amicable aid with selected models and aid generation methods, but it will be desirable to perform large-scale benchmarking studies with more diverse models, datasets, and generation methods \cite{dong20benchmarking}.

\bibliographystyle{IEEEbib}
\bibliography{refs}

\begin{thebibliography}{10}

\bibitem{su2018robustness}
Dong Su, Huan Zhang, Hongge Chen, Jinfeng Yi, Pin-Yu Chen, and Yupeng Gao,
\newblock ``Is robustness the cost of accuracy? -- {A} comprehensive study on
  the robustness of 18 deep image classification models,''
\newblock in {\em Proc. ECCV}, 2018.

\bibitem{goodfellow2015explaining}
Ian~J. Goodfellow, Jonathon Shlens, and Christian Szegedy,
\newblock ``Explaining and harnessing adversarial examples,''
\newblock in {\em Proc. ICLR}, 2015.

\bibitem{hwang2021just}
Jaehui Hwang, Jun-Hyuk Kim, Jun-Ho Choi, and Jong-Seok Lee,
\newblock ``Just one moment: {S}tructural vulnerability of deep action
  recognition against one frame attack,''
\newblock in {\em Proc. ICCV}, 2021.

\bibitem{simonyan2015very}
Karen Simonyan and Andrew Zisserman,
\newblock ``Very deep convolutional networks for large-scale image
  recognition,''
\newblock in {\em Proc. ICLR}, 2015.

\bibitem{szegedy2013intriguing}
Christian Szegedy, Wojciech Zaremba, Ilya Sutskever, Joan Bruna, Dumitru Erhan,
  Ian Goodfellow, and Rob Fergus,
\newblock ``Intriguing properties of neural networks,''
\newblock in {\em Proc. ICLR}, 2014.

\bibitem{kurakin2016adversarial}
Alexey Kurakin, Ian Goodfellow, and Samy Bengio,
\newblock ``Adversarial machine learning at scale,''
\newblock in {\em Proc. ICLR}, 2017.

\bibitem{carlini2017evaluating}
Nicholas Carlini and David Wagner,
\newblock ``Towards evaluating the robustness of neural networks,''
\newblock in {\em Proc. IEEE Symposium on Security and Privacy}, 2017.

\bibitem{luo2018towards}
Bo~Luo, Yannan Liu, Lingxiao Wei, and Qiang Xu,
\newblock ``Towards imperceptible and robust adversarial example attacks
  against neural networks,''
\newblock in {\em Proc. AAAI}, 2018.

\bibitem{madry2017towards}
Aleksander Madry, Aleksandar Makelov, Ludwig Schmidt, Dimitris Tsipras, and
  Adrian Vladu,
\newblock ``Towards deep learning models resistant to adversarial attacks,''
\newblock in {\em Proc. ICLR}, 2017.

\bibitem{tramer2018ensemble}
Florian Tramèr, Alexey Kurakin, Nicolas Papernot, Ian Goodfellow, Dan Boneh,
  and Patrick McDaniel,
\newblock ``Ensemble adversarial training: Attacks and defenses,''
\newblock in {\em Proc. ICLR}, 2018.

\bibitem{modas2019sparsefool}
Apostolos Modas, Seyed-Mohsen Moosavi-Dezfooli, and Pascal Frossard,
\newblock ``Sparse{F}ool: a few pixels make a big difference,''
\newblock in {\em Proc. CVPR}, 2019.

\bibitem{wong2020fast}
Eric Wong, Leslie Rice, and J~Zico Kolter,
\newblock ``Fast is better than free: Revisiting adversarial training,''
\newblock in {\em Proc. ICLR}, 2020.

\bibitem{croce2020reliable}
Francesco Croce and Matthias Hein,
\newblock ``Reliable evaluation of adversarial robustness with an ensemble of
  diverse parameter-free attacks,''
\newblock in {\em Proc. ICML}, 2020.

\bibitem{moosavi2017universal}
Seyed-Mohsen Moosavi-Dezfooli, Alhussein Fawzi, Omar Fawzi, and Pascal
  Frossard,
\newblock ``Universal adversarial perturbations,''
\newblock in {\em Proc. CVPR}, 2017.

\bibitem{stutz2019disentangling}
David Stutz, Matthias Hein, and Bernt Schiele,
\newblock ``Disentangling adversarial robustness and generalization,''
\newblock in {\em Proc. CVPR}, 2019.

\bibitem{song18constructing}
Yang Song, Rui Shu, Nate Kushman, and Stefano Ermon,
\newblock ``Constructing unrestricted adversarial examples with generative
  models,''
\newblock in {\em Proc. NeurIPS}, 2018.

\bibitem{tanay2016boundary}
Thomas Tanay and Lewis Griffin,
\newblock ``A boundary tilting perspective on the phenomenon of adversarial
  examples,''
\newblock {\em arXiv:1608.07690}, 2016.

\bibitem{lukas2019deep}
Nils Lukas, Yuxuan Zhang, and Florian Kerschbaum,
\newblock ``Deep neural network fingerprinting by conferrable adversarial
  examples,''
\newblock in {\em Proc. ICLR}, 2021.

\bibitem{xie2020adversarial}
Cihang Xie, Mingxing Tan, Boqing Gong, Jiang Wang, Alan~L. Yuille, and Quoc~V.
  Le,
\newblock ``Adversarial examples improve image recognition,''
\newblock in {\em Proc. CVPR}, 2020.

\bibitem{jalwana2020attack}
Mohammad A. A.~K. Jalwana, Naveed Akhtar, Mohammed Bennamoun, and Ajmal Mian,
\newblock ``Attack to explain deep representation,''
\newblock in {\em Proc. CVPR}, 2020.

\bibitem{deep2015nguyen}
Anh Nguyen, Jason Yosinski, and Jeff Clune,
\newblock ``Deep neural networks are easily fooled: High confidence predictions
  for unrecognizable images,''
\newblock in {\em Proc. CVPR}, 2015.

\bibitem{InkawhichLWICC20}
Nathan Inkawhich, Kevin~J. Liang, Binghui Wang, Matthew Inkawhich, Lawrence
  Carin, and Yiran Chen,
\newblock ``Perturbing across the feature hierarchy to improve standard and
  strict blackbox attack transferability,''
\newblock in {\em Proc. NeurIPS}, 2020.

\bibitem{zhao2021on}
Zhengyu Zhao, Zhuoran Liu, and Martha Larson,
\newblock ``On success and simplicity: A second look at transferable targeted
  attacks,''
\newblock in {\em Proc. NeurIPS}, 2021.

\bibitem{huang2021unlearnable}
Hanxun Huang, Xingjun Ma, Sarah~Monazam Erfani, James Bailey, and Yisen Wang,
\newblock ``Unlearnable examples: Making personal data unexploitable,''
\newblock in {\em Proc. ICLR}, 2021.

\bibitem{Pestana_2021_CVPR}
Camilo Pestana, Wei Liu, David Glance, Robyn Owens, and Ajmal Mian,
\newblock ``Assistive signals for deep neural network classifiers,''
\newblock in {\em Proc. CVPR Workshops}, 2021.

\bibitem{salman2021unadversarial}
Hadi Salman, Andrew Ilyas, Logan Engstrom, Sai Vemprala, Aleksander Madry, and
  Ashish Kapoor,
\newblock ``Unadversarial examples: Designing objects for robust vision,''
\newblock in {\em Proc. NeurIPS}, 2021.

\bibitem{krizhevsky2009learning}
Alex Krizhevsky and Geoffrey Hinton,
\newblock ``Learning multiple layers of features from tiny images,''
\newblock Tech. {R}ep., University of Toronto, Toronto, Ontario, 2009.

\bibitem{ILSVRC15}
Olga Russakovsky, Jia Deng, Hao Su, Jonathan Krause, Sanjeev Satheesh, Sean Ma,
  Zhiheng Huang, Andrej Karpathy, Aditya Khosla, Michael Bernstein,
  Alexander~C. Berg, and Li~Fei-Fei,
\newblock ``{ImageNet Large Scale Visual Recognition Challenge},''
\newblock {\em IJCV}, 2015.

\bibitem{he2016deep}
Kaiming He, Xiangyu Zhang, Shaoqing Ren, and Jian Sun,
\newblock ``Deep residual learning for image recognition,''
\newblock in {\em Proc. CVPR}, 2016.

\bibitem{sandler2018mobilenetv2}
Mark Sandler, Andrew Howard, Menglong Zhu, Andrey Zhmoginov, and Liang-Chieh
  Chen,
\newblock ``{MobileNetV2}: Inverted residuals and linear bottlenecks,''
\newblock in {\em Proc. CVPR}, 2018.

\bibitem{Li2020DefenseAA}
Yueru Li, Shuyu Cheng, Hang Su, and Jun Zhu,
\newblock ``Defense against adversarial attacks via controlling gradient
  leaking on embedded manifolds,''
\newblock in {\em Proc. ECCV}, 2020.

\bibitem{shamsabadi20colorfool}
Ali~Shahin Shamsabadi, Ricardo Sanchez-Matilla, and Andrea Cavallaro,
\newblock ``{ColorFool}: {S}emantic adversarial colorization,''
\newblock in {\em Proc. CVPR}, 2020.

\bibitem{zhao20towards}
Zhengyu Zhao, Zhuoran Liu, and Martha Larson,
\newblock ``Towards large yet imperceptible adversarial image perturbations
  with perceptual color distance,''
\newblock in {\em Proc. CVPR}, 2020.

\bibitem{bai21targeted}
Jiawang Bai, Baoyuan Wu, Yong Zhang, Yiming Li, Zhifeng Li, and Shu-Tao Xia,
\newblock ``Targeted attack against deep neural networks via flipping limited
  weight bits,''
\newblock in {\em Proc. ICLR}, 2021.

\bibitem{dong20benchmarking}
Yinpeng Dong, Qi-An Fu, Xiao Yang, Tianyu Pang, Hang Su, Zihao Xiao, and Jun
  Zhu,
\newblock ``Benchmarking adversarial robustness on image classification,''
\newblock in {\em Proc. CVPR}, 2020.

\end{thebibliography}

\end{document}